\documentclass[journal]{IEEEtran}

\usepackage{cite}
\usepackage{amsmath,amssymb,bm}
\usepackage{graphicx}
\usepackage{booktabs}
\usepackage{multirow}
\usepackage{placeins}
\usepackage{xcolor}
\usepackage{url}
\usepackage[breaklinks=true,colorlinks=true,linkcolor=black,citecolor=blue,urlcolor=black]{hyperref}
\newcommand{\best}[1]{\textcolor{red}{\textbf{#1}}}
\newcommand{\second}[1]{\underline{#1}}

\title{RPBA-Net: An Interpretable Residual Pyramid Bilateral Affine Network for RAW-Domain ISP Enhancement}
\author{
Yucheng~Xin,
Wu~Chen,
Xiang~Chen,
Guangwei~Gao,
Xinchun~Wang,
Ruize~Wu,
Dianjie~Lu,
Guijuan~Zhang,
Linwei~Fan, and
Zhuoran~Zheng$^{*}$
}

\begin{document}
\sloppy
\maketitle

\begingroup
\renewcommand{\thefootnote}{}
\footnotetext{\raggedright
Yucheng Xin, Xinchun Wang, Ruize Wu, Dianjie Lu, and Guijuan Zhang are with the School of Information Science and Engineering, Shandong Normal University, Jinan, China
(e-mail: 202411000525@stu.sdnu.edu.cn; 1198383781@qq.com; 202411000136@stu.sdnu.edu.cn; ludianjie@sdnu.edu.cn; zhangguijuan@sdnu.edu.cn).

Wu Chen is with the National University of Defense Technology, Changsha, China
(e-mail: wuchen5X@mail.ustc.edu.cn).

Xiang Chen and Guangwei Gao are with Nanjing University of Science and Technology, Nanjing, China
(e-mail: chenxiang@njust.edu.cn; gwgao@njust.edu.cn).

Linwei Fan is with Shandong University of Finance and Economics, Jinan, China
(e-mail: lwfan129@163.com).

Zhuoran Zheng is with Qilu University of Technology, Jinan, China
(e-mail: zhengzr@njust.edu.cn).

\textsuperscript{*}Corresponding author: Zhuoran Zheng.
}
\endgroup

\begin{abstract}
To address module fragmentation, uninterpretable mappings, and deployment constraints in RAW-domain demosaicing, color correction, and detail enhancement, this paper proposes RPBA-Net, an interpretable residual pyramid bilateral affine network for RAW-domain ISP enhancement. Given packed RAW as input, the method performs residual affine base reconstruction by estimating a base RGB representation and learning identity-guided residual affine corrections, thereby unifying demosaicing and enhancement. It further builds pyramid bilateral affine grids and combines guide-driven autoregressive adaptive slicing with adaptive cross-layer fusion to hierarchically model global tone restoration and local texture enhancement. In addition, smoothness, cross-scale consistency, and magnitude regularization terms are introduced to improve model stability, controllability, and structural interpretability. Extensive experiments demonstrate that RPBA-Net surpasses representative RAW-to-sRGB methods and achieves state-of-the-art performance in reconstruction fidelity and perceptual quality, while maintaining low model complexity and strong deployment potential for mobile and embedded platforms.
\end{abstract}


\section{Introduction}
With the development of mobile imaging and computational photography~\cite{10.1145/2980179.2980254}, ISP has become increasingly important in imaging pipelines~\cite{8478390}, and is now a key component of modern imaging workflows. Compared with RGB images, RAW data is closer to the original sensor response and preserves richer luminance, color, and detail information~\cite{Brooks_2019_CVPR,Xing_2021_CVPR,Kim_2024_CVPR}, making it more suitable for tasks such as demosaicing, denoising, color correction, and image enhancement. Existing RAW-domain imaging and learned ISP studies~\cite{10.1145/2980179.2980254,Chen_2018_CVPR,8478390} have also shown that direct modeling in the sensor domain helps reduce information loss and error accumulation, thereby improving the potential for high-quality imaging.

\begin{figure}[!t]
\centering
\includegraphics[width=\columnwidth]{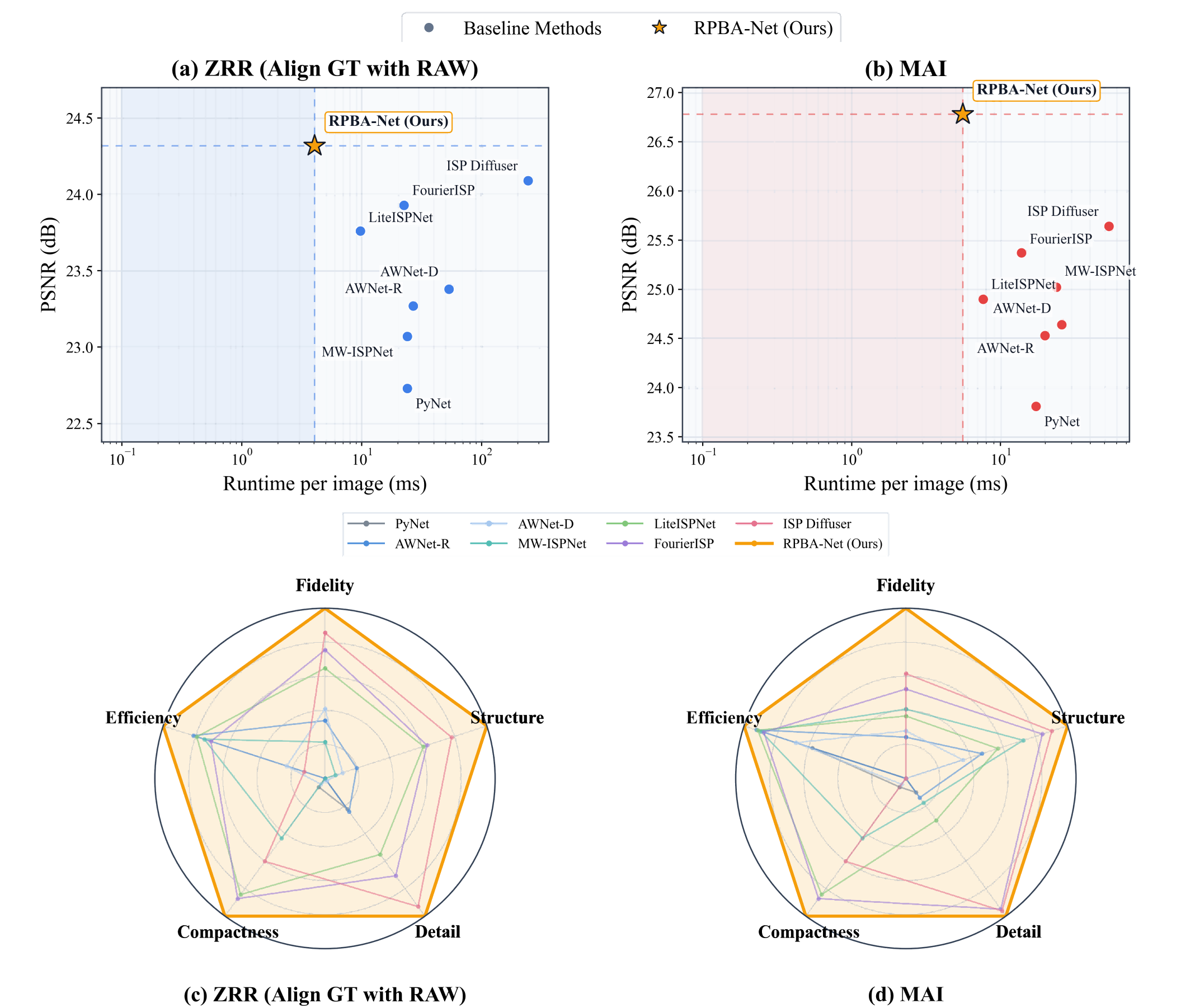}
\vspace{-1mm}
\caption{Overall comparison with representative RAW-to-sRGB methods on ZRR (Align GT with RAW) and MAI. RPBA-Net shows a favorable balance among reconstruction quality, inference efficiency, and model compactness.}
\label{fig:runtime_quality}
\end{figure}

Traditional ISP pipelines~\cite{8478390,Brooks_2019_CVPR,Zhang_2021_ICCV} generally consist of several independent modules, such as demosaicing, white balance, color correction, denoising, and sharpening. Although such methods are controllable and mature from an engineering perspective, interactions among modules are limited, and parameter design often depends heavily on manual experience and stage-wise tuning~\cite{8478390}. Consequently, compared with end-to-end learned pipelines~\cite{Ignatov_2020_CVPR_Workshops}, such stage-wise designs are less capable of globally optimizing the mapping from input RAW data to target RGB images.

A central goal of this work is to improve the trade-off between reconstruction quality and deployability in RAW-domain ISP enhancement. As shown in Fig.~\ref{fig:runtime_quality}, RPBA-Net achieves a more favorable balance among reconstruction quality, inference efficiency, and model compactness than representative baseline methods. In challenging scenarios with complex illumination, rich textures, or cross-device data settings, the errors introduced by upstream modules tend to propagate and accumulate~\cite{8478390,Zamir2020CycleISP,Zhang_2021_ICCV}, which degrades subsequent color restoration and detail enhancement and ultimately limits image quality.

In recent years, deep learning-based end-to-end RAW-to-RGB approaches have provided a new direction for ISP modeling. DeepISP~\cite{8478390} and PyNet~\cite{Ignatov_2020_CVPR_Workshops} first demonstrated the feasibility of learning a unified mapping from RAW to RGB. W-Net~\cite{9022089} and the method of Zhang et al.~\cite{Zhang_2021_ICCV} further improved RAW-to-sRGB learning under misaligned supervision. CycleISP~\cite{Zamir2020CycleISP}, InvISP~\cite{Xing_2021_CVPR}, ParamISP~\cite{Kim_2024_CVPR}, FourierISP~\cite{He_2024_FourierISP}, and RMFA-Net~\cite{li2024rmfanetneuralispreal} have advanced this direction from the perspectives of data synthesis, invertible modeling, camera-parameter-aware modeling, frequency-domain decoupling, and real RAW reconstruction, respectively. Meanwhile, studies targeting mobile and edge deployment indicate that learned ISP is moving from pursuing reconstruction quality alone toward jointly optimizing quality and efficiency. Attention mechanisms, architectural refinement, and lightweight design have therefore been increasingly adopted to reduce inference cost, as exemplified by AWNet~\cite{10.1007/978-3-030-67070-2_11}, PyNET-CA~\cite{Kim_2020}, the MAI 2021 Challenge~\cite{Ignatov_2021_CVPR}, CSANet~\cite{Hsyu_2021_CVPR}, LAN~\cite{Raimundo_2022_CVPR}, LW-ISP~\cite{Chen2022LWISPAL}, and MetaISP~\cite{ijcai2024p76}. However, most existing methods still rely on deep convolutional features or complex feature transformations~\cite{8478390,Ignatov_2020_CVPR_Workshops} to implicitly accomplish color recovery and detail enhancement, and therefore lack clear structural interpretability in their mapping process. At the same time, AWNet~\cite{10.1007/978-3-030-67070-2_11}, LAN~\cite{Raimundo_2022_CVPR}, and MetaISP~\cite{ijcai2024p76} also suggest that single-scale or black-box mappings often struggle to balance global tone restoration, local texture enhancement, and computational cost under high-resolution settings.

In contrast, bilateral grid representations~\cite{10.1145/1276377.1276497,10.1145/2980179.2982423,10.1145/3072959.3073592,rezaee2021imageenhancementbilaterallearning} jointly encode spatial positions and guide dimensions and can represent content-aware local color transformations in a structured manner. In particular, local affine mappings based on bilateral grids~\cite{10.1145/2980179.2982423,10.1145/3072959.3073592} naturally correspond to pixel-wise color transformation, while remaining compact, easy to visualize, and deployment-friendly~\cite{rezaee2021imageenhancementbilaterallearning}. This makes them highly suitable for RAW-domain ISP enhancement. Nevertheless, existing bilateral representations are mainly designed for RGB image enhancement or single-layer local mappings~\cite{10.1145/2980179.2982423,10.1145/3072959.3073592,rezaee2021imageenhancementbilaterallearning}, and thus cannot fully cover the hierarchical mapping process from global low-frequency tone adjustment to local high-frequency detail restoration. Moreover, most existing bilateral slicing schemes rely on fixed interpolation kernels~\cite{10.1145/2980179.2982423,10.1145/3072959.3073592}, which limit content-adaptive coefficient retrieval and lack explicit modeling of coarse-to-fine recursive dependencies. Therefore, constructing a bilateral mapping framework that jointly offers multi-scale representation capability, adaptive coefficient slicing, structural interpretability, and lightweight deployment remains a key challenge for RAW-domain ISP enhancement.

To address these issues, we propose RPBA-Net, an interpretable residual pyramid bilateral affine network for RAW-domain ISP enhancement. Given packed RAW data as input, the proposed method first performs residual affine base reconstruction, where a lightweight backbone estimates a base RGB representation and identity-guided residual affine corrections are learned on top of it~\cite{8478390,Ignatov_2020_CVPR_Workshops,10.1145/2980179.2982423,10.1145/3072959.3073592}, thereby unifying base demosaicing and subsequent enhancement within a single end-to-end framework. Furthermore, pyramid bilateral affine grids are constructed from multi-scale encoder features~\cite{Ignatov_2020_CVPR_Workshops,10.1145/1276377.1276497,10.1145/2980179.2982423,10.1145/3072959.3073592}, and a guide map~\cite{10.1145/2980179.2982423,10.1145/3072959.3073592} predicted from the base RGB image is combined with autoregressive adaptive slicing to recover pixel-wise affine coefficients through learned local interpolation and coarse-to-fine recursive refinement, thereby enabling structured modeling of hierarchical mappings from global tone restoration to local texture enhancement. To improve the cooperation among different mapping levels, the network also introduces a softmax-based adaptive cross-layer fusion module and further enhances stability and controllability through grid smoothness, cross-scale consistency, and magnitude regularization. In this way, RPBA-Net preserves structural interpretability~\cite{10.1145/2980179.2982423,10.1145/3072959.3073592} while also maintaining a lightweight design and deployment potential~\cite{Ignatov_2021_CVPR,Raimundo_2022_CVPR,ijcai2024p76}, providing a unified modeling scheme for RAW-domain ISP enhancement that balances reconstruction quality and deployability.

Our main contributions are summarized as follows:
\begin{enumerate}
\item We propose RPBA-Net, an interpretable residual pyramid bilateral affine network for RAW-domain ISP enhancement, which takes packed RAW data as input and jointly models base demosaicing, color mapping, and detail enhancement within a unified end-to-end framework.
\item We design pyramid bilateral affine grids with autoregressive adaptive slicing and adaptive cross-layer fusion, enabling hierarchical tone restoration and texture enhancement, while regularization improves model stability, controllability, and structural interpretability.
\item Extensive experiments demonstrate the effectiveness of the proposed method. The results show that RPBA-Net outperforms representative competing methods in reconstruction fidelity and perceptual quality while maintaining low model complexity.
\end{enumerate}

\section{Related Work}
\subsection{End-to-End RAW-to-RGB Methods}
The limitations of traditional ISP pipelines, such as stage separation, hand-crafted parameter tuning, and progressive error accumulation, have motivated researchers to explore end-to-end RAW-to-RGB learning frameworks. DeepISP~\cite{8478390} first demonstrated the feasibility of unified ISP modeling, while PyNet~\cite{Ignatov_2020_CVPR_Workshops} improved end-to-end RAW-to-RGB reconstruction through a pyramid architecture. W-Net~\cite{9022089} and the joint alignment-and-mapping method of Zhang et al.~\cite{Zhang_2021_ICCV} further alleviated the issue of inaccurately aligned RAW-RGB supervision. CycleISP~\cite{Zamir2020CycleISP} and InvISP~\cite{Xing_2021_CVPR} extended this line of research by jointly modeling forward and inverse ISP processes. More recently, FourierISP~\cite{He_2024_FourierISP}, ParamISP~\cite{Kim_2024_CVPR}, and RMFA-Net~\cite{li2024rmfanetneuralispreal} have further improved mapping quality and generalization from the perspectives of frequency-domain decoupling, camera-parameter modeling, and real RAW reconstruction, respectively. However, most of these methods still rely on deep convolutional features to implicitly regress the final image. As a result, the color recovery and detail enhancement processes lack explicit structural representation, and local correction coefficients are often hidden in black-box mappings, leaving room for improvement in interpretability and mapping controllability.

\subsection{Lightweight Neural ISP Methods}
As learning-based ISP moves toward practical deployment, reducing model complexity and inference cost while preserving reconstruction quality has become an important research direction. Existing studies mainly improve deployability through efficient network structures, attention mechanisms, and model compression strategies. AWNet~\cite{10.1007/978-3-030-67070-2_11} and PyNET-CA~\cite{Kim_2020} improve mobile RAW-to-RGB reconstruction by introducing attention and architectural refinements, while the MAI 2021 Challenge~\cite{Ignatov_2021_CVPR} further promoted efficient learned ISP for NPU deployment. Subsequently, CSANet~\cite{Hsyu_2021_CVPR}, Del-Net~\cite{gupta2021delnetsinglestagenetworkmobile}, LAN~\cite{Raimundo_2022_CVPR}, and LW-ISP~\cite{Chen2022LWISPAL} further optimize model complexity and inference efficiency from the perspectives of channel-spatial attention, single-stage multi-scale design, lightweight attention networks, and distillation-based compression, respectively. More recently, MetaISP~\cite{ijcai2024p76} has further shown that learned ISP can still achieve strong reconstruction quality under tight parameter and FLOP budgets. Overall, these methods mainly focus on optimizing model complexity and inference efficiency, providing important references for the practical deployment of learning-based ISP.

\subsection{Bilateral Grid Representations}
Compared with directly predicting the final image, bilateral representations provide a more structured parameterization for image enhancement. JBU~\cite{10.1145/1276377.1276497} established an early foundation for high-resolution guided recovery. BGU~\cite{10.1145/2980179.2982423} further fits local affine models in bilateral space and transfers them from low-resolution pairs to high-resolution images, demonstrating the efficiency of bilateral representations for color transformation and image enhancement. Based on this idea, HDRNet~\cite{10.1145/3072959.3073592} combines deep learning with bilateral grids and predicts local affine transformation coefficients through guide-based slicing, achieving high-quality enhancement with low computational cost. Later, the bilateral learning method of Rezaee et al.~\cite{rezaee2021imageenhancementbilaterallearning} further extended this framework for image enhancement. Compared with direct regression methods, bilateral grids can represent content-aware local color transformations with compact parameters and also offer good visualization and analysis capability, making them well suited for pixel-wise color adjustment in ISP tasks. Overall, existing studies have demonstrated the advantages of bilateral grid representations for efficient image enhancement and laid an important foundation for subsequent structured ISP modeling.

\section{Method}

\subsection{Overview}
\begin{figure*}[!t]
\centering
\includegraphics[width=\textwidth]{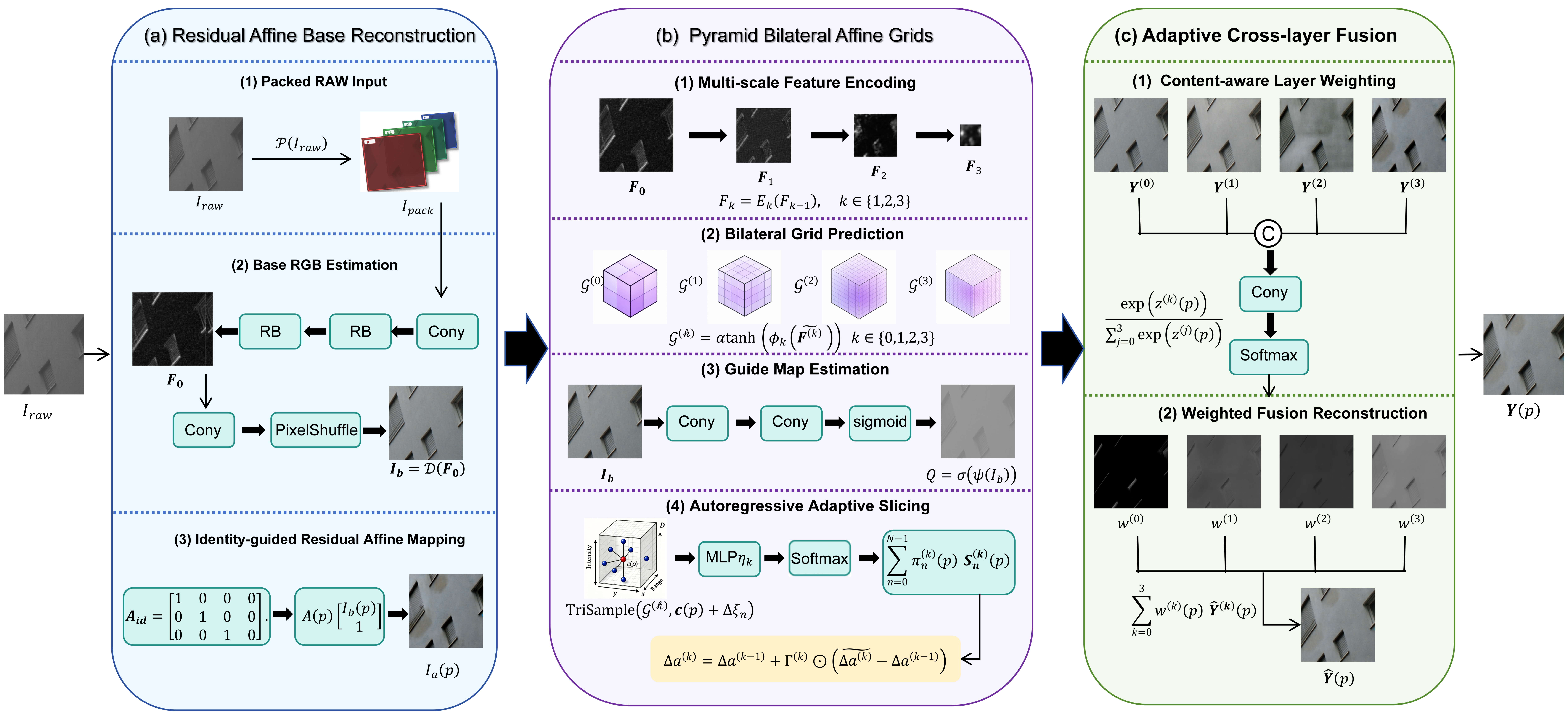}
\caption{Overall framework of RPBA-Net. The network performs residual affine base reconstruction, predicts pyramid bilateral affine grids, and adaptively fuses multi-scale affine outputs for RAW-domain ISP enhancement.}
\label{fig:method_framework}
\end{figure*}

RPBA-Net is an interpretable residual pyramid bilateral affine network for Packed RAW-to-RGB reconstruction~\cite{Chen_2018_CVPR,8478390,Ignatov_2020_CVPR_Workshops}, designed to jointly accomplish base demosaicing, color mapping, and detail enhancement within a unified end-to-end framework. Given a sensor RAW image $\bm{I}_{\mathrm{raw}}\in\mathbb{R}^{2H\times2W}$, the network first forms a packed RAW representation $\bm{I}_{\mathrm{pack}}=\mathcal{P}(\bm{I}_{\mathrm{raw}})\in\mathbb{R}^{4\times H\times W}$ and outputs the reconstructed RGB image $\hat{\bm{Y}}\in\mathbb{R}^{3\times 2H\times 2W}$. Overall, RPBA-Net consists of three key parts. First, a lightweight backbone extracts base features and produces an initial RGB representation, while pixel-wise residual color affine correction is learned using ideas from residual learning~\cite{He_2016_CVPR} and local affine mapping models~\cite{10.1145/2980179.2982423,10.1145/3072959.3073592}. Second, multi-scale features are used to construct four 3D bilateral affine grids~\cite{10.1145/1276377.1276497,10.1145/2980179.2982423,10.1145/3072959.3073592}, and guide-driven autoregressive adaptive slicing~\cite{10.1145/2980179.2982423,10.1145/3072959.3073592} generates pixel-wise affine coefficients for hierarchical modeling from global tone restoration to local texture enhancement. Third, an adaptive cross-layer fusion stage performs content-aware weighting over the four branches, while smoothness, consistency, and magnitude constraints improve overall reconstruction stability and structural controllability. The overall framework is illustrated in Fig.~\ref{fig:method_framework}.

\subsection{Residual Affine Base Reconstruction}
To make the Packed RAW-to-RGB mapping~\cite{Chen_2018_CVPR,8478390,Ignatov_2020_CVPR_Workshops} structurally interpretable, we first introduce residual affine base reconstruction. This stage does not aim to generate the final result in a single step. Instead, it first produces a stable base RGB representation and then unifies subsequent enhancement as pixel-wise residual affine correction over this base image.

\subsubsection{Packed RAW Input}
The sensor RAW data is typically arranged as a Bayer mosaic~\cite{Brooks_2019_CVPR,Chen_2018_CVPR}, which is not well suited for efficient convolutional modeling in its original form. We therefore convert the Bayer RAW image into a four-channel packed representation~\cite{Chen_2018_CVPR,Zhang_2021_ICCV}, where each channel corresponds to one sampling position. This operation preserves the original sensor response while making the input more suitable for feature extraction.

The packing process is written as
\begin{equation}
\begin{gathered}
\bm{I}_{\mathrm{pack}}=\mathcal{P}(\bm{I}_{\mathrm{raw}})
=[\bm{R},\bm{G}_{1},\bm{G}_{2},\bm{B}],\\
\bm{I}_{\mathrm{pack}}(:,i,j)=
\begin{bmatrix}
\bm{I}_{\mathrm{raw}}(2i,2j)\\
\bm{I}_{\mathrm{raw}}(2i,2j+1)\\
\bm{I}_{\mathrm{raw}}(2i+1,2j)\\
\bm{I}_{\mathrm{raw}}(2i+1,2j+1)
\end{bmatrix},
\end{gathered}
\end{equation}
where $\bm{I}_{\mathrm{raw}}\in\mathbb{R}^{2H\times2W}$ denotes the Bayer mosaic RAW image, $\mathcal{P}(\cdot)$ is the packing operator, $\bm{I}_{\mathrm{pack}}\in\mathbb{R}^{4\times H\times W}$ is the packed RAW representation, and $i=0,\ldots,H-1$, $j=0,\ldots,W-1$. Under the default RGGB layout, this indexing rule means that each packed location stores the four sensor responses of the corresponding $2\times2$ mosaic cell. Other Bayer layouts are first sampled according to their own CFA order and then reorganized to the same channel convention $[\bm{R},\bm{G}_{1},\bm{G}_{2},\bm{B}]$. Therefore, packing only transfers the local CFA neighborhood from the spatial domain to the channel domain, enabling efficient low-resolution processing while retaining the local structure required for color reconstruction.

\subsubsection{Base RGB Estimation}
After obtaining the packed RAW representation, a lightweight backbone is applied to extract base features and generate an initial RGB result. This branch is responsible for base demosaicing and preliminary color recovery, providing a stable starting point for subsequent affine enhancement.

Formally, the base feature extraction and base RGB estimation are defined as
\begin{equation}
\begin{gathered}
\bm{F}_{0}=\mathcal{E}_{0}(\bm{I}_{\mathrm{pack}})
=\mathrm{RB}_{2}\!\left(\mathrm{RB}_{1}\!\left(\mathrm{Conv}_{3\times3}^{4\rightarrow32}(\bm{I}_{\mathrm{pack}})\right)\right),\\
\bm{I}_{b}=\mathcal{D}(\bm{F}_{0})
=\mathrm{PS}_{2}\!\left(\mathrm{Conv}_{3\times3}^{32\rightarrow12}(\bm{F}_{0})\right),
\end{gathered}
\end{equation}
where $\mathcal{E}_{0}(\cdot)$ denotes the lightweight backbone encoder, $\bm{F}_{0}$ is the base feature map, $\mathcal{D}(\cdot)$ denotes the demosaicing and upsampling decoder, and $\bm{I}_{b}\in\mathbb{R}^{3\times 2H\times 2W}$ is the base RGB representation. $\mathcal{E}_{0}(\cdot)$ uses one $3\times3$ convolution and two residual blocks~\cite{He_2016_CVPR} while preserving the packed-domain resolution. $\mathcal{D}(\cdot)$ first predicts 12 channels and then applies the sub-pixel rearrangement $\mathrm{PS}_{2}(\cdot)$~\cite{Shi_2016_CVPR} to jointly perform demosaicing and $\times2$ upsampling. The resulting $\bm{I}_{b}$ serves as the reference basis for the subsequent residual affine modeling.

\subsubsection{Identity-guided Residual Affine Mapping}
On top of the base RGB representation, we further introduce a pixel-wise color affine transformation~\cite{10.1145/2980179.2982423,10.1145/3072959.3073592,rezaee2021imageenhancementbilaterallearning} to unify color correction and detail enhancement. Directly regressing absolute affine coefficients may lead to overly large mapping amplitudes, unstable optimization, and color drift. To address this issue, we impose an identity-mapping prior and adopt residual parameterization~\cite{He_2016_CVPR}, so that the network only learns incremental corrections over the base RGB image.

For an arbitrary pixel location $p\in\Omega$, where $\Omega=\{1,\ldots,2H\}\times\{1,\ldots,2W\}$ denotes the output RGB pixel domain, the residual affine correction is parameterized by a 12-D vector:
\begin{equation}
\Delta\bm{a}(p)=\left[\delta a_{1}(p),\delta a_{2}(p),\ldots,\delta a_{12}(p)\right]^{\top},
\end{equation}
where $\Delta\bm{a}(p)\in\mathbb{R}^{12}$ stores the residual parameters of a $3\times 4$ color affine transform. To illustrate the residual affine correction at a single pixel, we first write its generic form as
\begin{equation}
\bm{A}(p)=\bm{A}_{id}+\operatorname{mat}(\Delta\bm{a}(p)), \qquad
\bm{A}_{id}=
\begin{bmatrix}
1 & 0 & 0 & 0 \\
0 & 1 & 0 & 0 \\
0 & 0 & 1 & 0
\end{bmatrix}.
\end{equation}
Here, $\operatorname{mat}(\cdot)$ reshapes the 12-D residual vector into a $3\times4$ matrix. For each pixel, $\bm{I}_{b}(p)\in\mathbb{R}^{3}$ denotes the base RGB column vector. The corrected RGB value at pixel $p$ is then computed as
\begin{equation}
\bm{I}_{a}(p)=\bm{A}(p)
\begin{bmatrix}
\bm{I}_{b}(p) \\
1
\end{bmatrix}.
\end{equation}
This generic single-pixel affine correction will later be instantiated at each bilateral-grid level as $\bm{A}^{(k)}(p)$ and $\hat{\bm{Y}}^{(k)}(p)$.
Under this formulation, the network does not regenerate the entire RGB image from scratch. Instead, it performs pixel-wise residual color correction around the base RGB image. In this way, base demosaicing and subsequent enhancement are unified within a single affine modeling framework~\cite{10.1145/2980179.2982423,10.1145/3072959.3073592}, while the color adjustment process remains explicitly interpretable.

\subsection{Pyramid Bilateral Affine Grids}
After establishing the residual affine formulation, we further model the generation of pixel-wise affine parameters from the perspectives of efficiency, stability, and hierarchy. Since a single-scale local mapping is usually insufficient to jointly handle global low-frequency tone recovery and local high-frequency texture enhancement, we introduce pyramid bilateral affine grids~\cite{10.1145/1276377.1276497,10.1145/2980179.2982423,10.1145/3072959.3073592} and construct four bilateral grids from multi-scale encoder features, so that different levels are responsible for color and detail adjustment at different granularities.

\subsubsection{Multi-scale Feature Encoding}
To provide appropriate context for bilateral grid prediction at different scales, we stack lightweight encoder layers on top of the base feature $\bm{F}_{0}$ and progressively extract multi-scale semantic and structural features. This process is expressed as
\begin{equation}
\begin{aligned}
\bm{F}_{1}&=\mathcal{E}_{1}(\bm{F}_{0})
=\mathrm{RB}\!\left(\mathrm{Conv}_{3\times3,s=2}^{32\rightarrow48}(\bm{F}_{0})\right),\\
\bm{F}_{2}&=\mathcal{E}_{2}(\bm{F}_{1})
=\mathrm{RB}\!\left(\mathrm{Conv}_{3\times3,s=2}^{48\rightarrow64}(\bm{F}_{1})\right),\\
\bm{F}_{3}&=\mathcal{E}_{3}(\bm{F}_{2})
=\mathrm{RB}\!\left(\mathrm{Conv}_{3\times3,s=2}^{64\rightarrow80}(\bm{F}_{2})\right),
\end{aligned}
\end{equation}
where $\bm{F}_{1}$, $\bm{F}_{2}$, and $\bm{F}_{3}$ denote hierarchical encoder features from shallow to deep levels. All three encoder stages follow the same pattern of stride-2 convolutional downsampling plus residual refinement~\cite{He_2016_CVPR}. Their output channels are 48, 64, and 80, and their spatial resolutions are reduced to $\tfrac{1}{2}$, $\tfrac{1}{4}$, and $\tfrac{1}{8}$ of $\bm{F}_{0}$, respectively. Shallow features emphasize local textures and edges, whereas deep features are more suitable for global tone and color modeling.

\subsubsection{Bilateral Grid Prediction}
Based on the multi-scale features, we construct four 3D bilateral affine grids~\cite{10.1145/2980179.2982423,10.1145/3072959.3073592,rezaee2021imageenhancementbilaterallearning} with different spatial resolutions to carry hierarchical residual color mappings. The grid levels are ordered from coarse to fine, with $\bm{F}^{(0)}=\bm{F}_{3}$, $\bm{F}^{(1)}=\bm{F}_{2}$, $\bm{F}^{(2)}=\bm{F}_{1}$, and $\bm{F}^{(3)}=\bm{F}_{0}$. Let $\mathcal{G}^{(k)}$ denote the bilateral grid at level $k$. Its prediction is formulated as
\begingroup
\small
\begin{equation}
\begin{gathered}
\tilde{\bm{F}}^{(k)}
=\mathrm{Resize}(\bm{F}^{(k)},S_{k},S_{k}),
\\
(S_{0},S_{1},S_{2},S_{3})=(16,32,64,128), \\[0.4ex]
\phi_{k}\!\left(\tilde{\bm{F}}^{(k)}\right)
=\mathrm{Conv}_{1\times1}\!\left(\bm{U}^{(k)}\right), \\
\bm{U}^{(k)}
=\mathrm{UpBlock}_{2}\!\left(\bm{V}^{(k)}\right), \\
\bm{V}^{(k)}
=\mathrm{ResBlock}_{2}\!\left(\bm{T}^{(k)}\right), \\
\bm{T}^{(k)}
=\mathrm{DownBlock}_{2}\!\left(
\mathrm{Conv}_{3\times3}\!\left(\tilde{\bm{F}}^{(k)}\right)\right), \\[0.4ex]
\mathcal{G}^{(k)}
=\alpha\tanh\!\big(\phi_{k}(\tilde{\bm{F}}^{(k)})\big),
\qquad
k\in\{0,1,2,3\},
\end{gathered}
\end{equation}
\endgroup
where $\phi_{k}(\cdot)$ denotes the lightweight grid prediction subnet at level $k$, $\alpha$ is a residual amplitude scaling factor, and $\tilde{\bm{F}}^{(k)}$ is the resized feature used for grid prediction. Each grid satisfies $\mathcal{G}^{(k)}\in\mathbb{R}^{12\times D\times S_{k}\times S_{k}}$, where $D$ is fixed and $S_{k}$ controls the spatial resolution. Thus, $\mathcal{G}^{(0)},\mathcal{G}^{(1)},\mathcal{G}^{(2)}$, and $\mathcal{G}^{(3)}$ are ordered from coarse to fine. $\phi_{k}(\cdot)$ uses a lightweight Tiny U-Net~\cite{10.1007/978-3-319-24574-4_28} to output $12D$ channels. All four levels share the same topology and differ only in input scale. The 12 channels correspond to the coefficients of an affine $3\times4$ transform, while the 3D bilateral structure jointly encodes channel parameters, guide depth, and spatial resolution.

\subsubsection{Guide Map Estimation}
To provide the depth coordinate for subsequent bilateral indexing, we predict a single-channel guide map~\cite{10.1145/2980179.2982423,10.1145/3072959.3073592} from the base RGB image. The guide map is generated as
\begin{equation}
\begin{gathered}
\bm{Q}=\sigma\!\left(\psi(\bm{I}_{b})\right),\\
\psi(\bm{I}_{b})=
\mathrm{Conv}_{1\times1}^{16\rightarrow1}\!\left(
\mathrm{ConvAct}_{3\times3}^{3\rightarrow16}(\bm{I}_{b})\right),
\end{gathered}
\end{equation}
where $\psi(\cdot)$ denotes the guide prediction branch, $\sigma(\cdot)$ is the sigmoid function, and $\bm{Q}\in[0,1]^{1\times 2H\times 2W}$ is the normalized guide map. $\psi(\cdot)$ consists of one $3\times3$ convolution and one $1\times1$ projection, and predicts a single-channel guide response from $\bm{I}_{b}$. The final sigmoid function normalizes it to $[0,1]$ and provides the depth coordinate for subsequent bilateral indexing.

\subsubsection{Autoregressive Adaptive Slicing}
Given the predicted guide map, we further perform autoregressive adaptive bilateral slicing over the four bilateral grids to recover content-aware pixel-wise affine mappings from the low-dimensional grid representation. Instead of relying on a fixed trilinear interpolation kernel~\cite{10.1145/2980179.2982423,10.1145/3072959.3073592}, each pixel aggregates a small neighborhood in bilateral space with learned weights, while the affine coefficient vector estimated at the previous coarser level is recursively injected as a prior. In this way, each pixel can retrieve the most suitable local affine coefficients according to its spatial location, image content, and cross-scale context. For pixel $p\in\Omega$, $x(p),y(p)\in[-1,1]$ denote its normalized spatial coordinates in the output image domain. For the $k$-th bilateral grid, we first sample a local neighborhood around $p$:
\begingroup
\small
\begin{equation}
\begin{gathered}
\bm{c}(p)=\left[x(p),\,y(p),\,2\bm{Q}(p)-1\right]^{\top},\\[0.8ex]
\bm{S}^{(k)}_{n}(p)=
\mathcal{S}_{\Delta\xi_{n}}\!\left(\mathcal{G}^{(k)},\bm{Q}(p)\right),\\
\bm{S}^{(k)}_{n}(p)=
\mathrm{TriSample}\!\left(\mathcal{G}^{(k)},\bm{c}(p)+\Delta\xi_{n}\right),\\[-0.2ex]
n\in\{0,1,\ldots,N-1\},
\end{gathered}
\end{equation}
\endgroup
where $N=7$ in our implementation, $\Delta\xi_{n}=(\Delta x_{n},\Delta y_{n},\Delta d_{n})$ enumerates the center location together with one-grid-step offsets along the horizontal, vertical, and guide-depth axes in the normalized bilateral coordinate system, and $\mathcal{S}_{\Delta\xi_{n}}(\cdot)$ denotes shifted bilateral slicing. For each pixel, the model first forms the bilateral coordinate $\bm{c}(p)$ from its spatial position and guide value~\cite{10.1145/1276377.1276497}. The shifted slicing result $\bm{S}^{(k)}_{n}(p)\in\mathbb{R}^{12}$ is then implemented by trilinear sampling~\cite{10.1145/2980179.2982423,10.1145/3072959.3073592} at the offset coordinate $\bm{c}(p)+\Delta\xi_{n}$. This uses the center point and six axis-aligned neighbors, yielding a 7-sample local bilateral neighborhood. Based on the center sample $\bm{S}^{(k)}_{0}(p)$ and the previous-scale residual vector $\Delta\bm{a}^{(k-1)}(p)$, the interpolation weights are predicted as
\begin{equation}
\begin{gathered}
\bm{\pi}^{(k)}(p)=\mathrm{Softmax}\!\left(\eta_{k}\!\left(\bm{h}^{(k)}(p)\right)\right),\\
\begin{aligned}
\bm{h}^{(k)}(p)&=[\bm{I}_{b}(p)\|\bm{Q}(p)\|\bm{S}^{(k)}_{0}(p)\|\Delta\bm{a}^{(k-1)}(p)],\\
\eta_{k}(\bm{u})&=\mathrm{Conv}_{1\times1}^{24\rightarrow N}\!\left(\mathrm{SiLU}\!\left(\mathrm{Conv}_{1\times1}^{28\rightarrow24}(\bm{u})\right)\right),
\end{aligned}
\end{gathered}
\end{equation}
where $\eta_{k}(\cdot)$ denotes a lightweight $1\times1$ mapping network, and $\Delta\bm{a}^{(-1)}(p)=\bm{0}$ for the coarsest level. It uses two $1\times1$ convolutions with hidden width 24, takes the 28-channel context as input, and outputs $N$ logits that are normalized by softmax into adaptive interpolation weights. The adaptively interpolated residual vector is then written as
\begin{equation}
\widetilde{\Delta\bm{a}}^{(k)}(p)=\sum_{n=0}^{N-1}\pi_{n}^{(k)}(p)\,\bm{S}^{(k)}_{n}(p).
\end{equation}
To further establish coarse-to-fine autoregression across levels, we predict a channel-wise update gate
\begin{equation}
\begin{gathered}
\bm{\Gamma}^{(k)}(p)
=\sigma\!\left(\rho_{k}\!\left(\bm{g}^{(k)}(p)\right)\right),\\
\begin{aligned}
\bm{g}^{(k)}(p)
&=[\bm{I}_{b}(p)\|\bm{Q}(p)\|\widetilde{\Delta\bm{a}}^{(k)}(p)
\|\Delta\bm{a}^{(k-1)}(p)],\\
\rho_{k}(\bm{v})
&=\mathrm{Conv}_{1\times1}^{24\rightarrow12}\!\left(
\mathrm{SiLU}\!\left(\mathrm{Conv}_{1\times1}^{28\rightarrow24}(\bm{v})\right)\right),
\end{aligned}\\[-0.2ex]
k\in\{1,2,3\},
\end{gathered}
\end{equation}
where $\rho_{k}(\cdot)$ is another two-layer $1\times1$ predictor whose output is passed through the sigmoid gate to produce a 12-channel update vector. This gate is used for the refinement levels $k\in\{1,2,3\}$ and estimates how much each affine-coefficient channel should absorb the newly sliced residual. The coarsest level is initialized directly and does not use a gate.

Using this gate, the residual affine vector is then recursively refined from the previous coarser estimate:
\begin{equation}
\begin{gathered}
\Delta\bm{a}^{(0)}(p)=\widetilde{\Delta\bm{a}}^{(0)}(p),\\
\resizebox{0.99\columnwidth}{!}{$\displaystyle
\Delta\bm{a}^{(k)}(p)=\Delta\bm{a}^{(k-1)}(p)+\bm{\Gamma}^{(k)}(p)\odot
\left(\widetilde{\Delta\bm{a}}^{(k)}(p)-\Delta\bm{a}^{(k-1)}(p)\right),$}\\[-0.2ex]
k\in\{1,2,3\},
\end{gathered}
\end{equation}
where $\Delta\bm{a}^{(0)}(p)$ is initialized by the first-level sliced residual, and each subsequent level $k\in\{1,2,3\}$ updates the previous estimate through a gated residual correction.
The corresponding affine matrix and candidate reconstructed output are
\begin{equation}
\begin{gathered}
\bm{A}^{(k)}(p)=\bm{A}_{id}+\operatorname{mat}(\Delta\bm{a}^{(k)}(p)),\\[0.4ex]
\hat{\bm{Y}}^{(k)}(p)=\bm{A}^{(k)}(p)
\begin{bmatrix}
\bm{I}_{b}(p) \\
1
\end{bmatrix}.
\end{gathered}
\end{equation}
Compared with fixed trilinear slicing~\cite{10.1145/2980179.2982423,10.1145/3072959.3073592}, the interpolation weights are now explicitly content-aware and the branch coefficients are recursively refined from coarse to fine. Coarse levels provide stable global priors, while fine levels adaptively deviate from those priors only where local edges or textures require additional correction. Through this procedure, each bilateral grid still produces an output with a clear interpretation, and the residual pyramid bilateral affine grids remain the core hierarchical representation of RPBA-Net.

\subsection{Adaptive Cross-layer Fusion}
After autoregressive adaptive slicing, the four bilateral grids provide complementary multi-scale mappings, but different image regions do not rely equally on all levels. We therefore introduce an adaptive cross-layer fusion module that directly takes the four candidate RGB outputs as input, predicts pixel-wise fusion weights with a lightweight mapping, and performs explicit weighted reconstruction. No additional black-box decoder is introduced, so the fusion remains in the image domain and the contribution of each scale stays directly interpretable.

\subsubsection{Content-aware Layer Weighting}
Given the four candidate outputs $\{\hat{\bm{Y}}^{(0)},\hat{\bm{Y}}^{(1)},\hat{\bm{Y}}^{(2)},\hat{\bm{Y}}^{(3)}\}$, where each candidate corresponds to an explicit color correction result produced by an autoregressively refined bilateral grid at a specific scale, we first organize them into a cross-layer fusion input by concatenating them along the channel dimension:
\begin{equation}
\bm{C}=[\hat{\bm{Y}}^{(0)} \| \hat{\bm{Y}}^{(1)} \| \hat{\bm{Y}}^{(2)} \| \hat{\bm{Y}}^{(3)}], \qquad
\bm{C}\in\mathbb{R}^{12\times 2H\times 2W}.
\end{equation}
This concatenated tensor gathers the RGB responses of all four levels at the same pixel location, allowing the fusion module to directly compare their reconstruction tendencies in the image domain. Based on $\bm{C}$, a lightweight $1\times 1$ convolution is used to estimate the cross-layer fusion responses:
\begin{equation}
\bm{Z}=\mathcal{F}_{1\times 1}(\bm{C})=\mathrm{Conv}_{1\times1}^{12\rightarrow4}(\bm{C}), \qquad
\bm{Z}\in\mathbb{R}^{4\times 2H\times 2W},
\end{equation}
where $\mathcal{F}_{1\times 1}(\cdot)$ denotes the $1\times1$ convolution and acts as a single-layer projection from 12 channels to 4 channels. Since it only mixes channels at each pixel, the fusion decision is made directly from the local responses of the four branches without introducing an additional spatial decoder. A softmax operation is then applied along the level/channel dimension to normalize the responses across levels:
\begin{equation}
\begin{gathered}
\bm{W}=\mathrm{Softmax}(\bm{Z}),\\
w^{(k)}(p)=
\frac{\exp(z^{(k)}(p))}
{\sum_{j=0}^{3}\exp(z^{(j)}(p))},
\qquad
\sum_{k=0}^{3} w^{(k)}(p)=1,
\end{gathered}
\end{equation}
where $\bm{W}\in\mathbb{R}^{4\times 2H\times 2W}$ represents the pixel-wise fusion weights of the four branches, and $w^{(k)}(p)$ is the softmax weight of level $k$ at pixel $p$. The four weights sum to 1, allowing the model to adaptively select suitable scales while suppressing uncontrolled accumulation, color shifts, and local over-enhancement.

\subsubsection{Weighted Fusion Reconstruction}
After obtaining the cross-layer weights, the final output is computed as the pixel-wise weighted sum of the four branch results. Since each candidate output $\hat{\bm{Y}}^{(k)}$ is reconstructed under a scale-specific bilateral affine mapping applied to the same base RGB image, the final fusion process can be interpreted as a content-aware selection and combination among four hierarchical reconstruction candidates:
\begin{equation}
\hat{\bm{Y}}(p)=\sum_{k=0}^{3} w^{(k)}(p)\,\hat{\bm{Y}}^{(k)}(p).
\end{equation}
This equation shows that the final result is formed by pixel-wise weighted contributions from all four levels rather than any single scale alone. Because the weights are normalized, $\hat{\bm{Y}}(p)$ is a convex combination of the four candidates, which improves numerical stability. Smooth regions typically rely more on coarse branches, whereas edges and textures rely more on fine branches. In this way, adaptive cross-layer fusion balances global tone consistency and local detail recovery.

\subsection{Loss Function}
The overall objective consists of a pixel reconstruction term and three grid regularization terms, jointly constraining the final reconstruction result and the structural stability of the bilateral grids. The reconstruction loss enforces consistency between the output image and the ground-truth RGB image, the smoothness term suppresses abrupt variations of the bilateral grids along spatial and depth dimensions, the cross-scale consistency term improves coordination between adjacent grid levels, and the magnitude term constrains the overall amplitude of the residual affine coefficients.

\subsubsection{Overall Objective}
Accordingly, the final training objective is formulated as
\begin{equation}
\mathcal{L}=\mathcal{L}_{rec}
+\lambda_{s}\mathcal{L}_{smooth}
+\lambda_{c}\mathcal{L}_{cons}
+\lambda_{m}\mathcal{L}_{mag},
\end{equation}
where $\lambda_{s}$, $\lambda_{c}$, and $\lambda_{m}$ are the weighting factors for the smoothness, cross-scale consistency, and magnitude regularization terms, respectively. The specific definitions of these loss terms are given below.

\subsubsection{Pixel Reconstruction Loss}
To directly supervise the difference between the final output and the target RGB image, we adopt an $L_{1}$ reconstruction loss:
\begin{equation}
\mathcal{L}_{rec}=\frac{1}{|\Omega|}\sum_{p\in\Omega}\left\|\hat{\bm{Y}}(p)-\bm{Y}(p)\right\|_{1},
\end{equation}
where $\hat{\bm{Y}}$ is the network prediction, $\bm{Y}$ is the ground-truth RGB image, and $\Omega$ denotes the pixel domain. The $L_{1}$ loss provides stable supervision for preserving both color and structural information. For the grid regularizers below, $\|\cdot\|_{1}$ denotes the mean absolute value over all elements of its tensor argument, so different grid resolutions are normalized by their element counts.

\subsubsection{Grid Smoothness Regularization}
To avoid discontinuous fluctuations of the bilateral grids along spatial and guide-depth dimensions, we impose a smoothness regularization on each level:
\begin{equation}
\resizebox{0.98\columnwidth}{!}{$\displaystyle
\mathcal{L}_{smooth}
=\frac{1}{4}\sum_{k=0}^{3}
\left(
\left\|\nabla_{x}\mathcal{G}^{(k)}\right\|_{1}
+\left\|\nabla_{y}\mathcal{G}^{(k)}\right\|_{1}
+\left\|\nabla_{d}\mathcal{G}^{(k)}\right\|_{1}
\right),$}
\end{equation}
where $\nabla_{x}$, $\nabla_{y}$, and $\nabla_{d}$ denote finite differences of the bilateral grid along the horizontal, vertical, and depth dimensions, respectively. This term encourages locally smooth coefficient variation and helps suppress color discontinuities and local artifacts.

\subsubsection{Cross-scale Consistency Regularization}
Since the four bilateral grids jointly serve the same affine reconstruction objective, we further introduce a cross-scale consistency loss to improve structural coordination between adjacent levels:
\begin{equation}
\mathcal{L}_{cons}=\frac{1}{3}\sum_{k=1}^{3}
\left\|
\mathcal{U}\!\left(\mathcal{G}^{(k-1)}\right)-\mathcal{G}^{(k)}
\right\|_{1},
\end{equation}
where $\mathcal{U}(\cdot)$ denotes trilinear upsampling of a coarse bilateral grid so that it matches the depth and spatial resolution of the adjacent finer grid under the coarse-to-fine order defined above. This constraint prevents neighboring levels from learning conflicting mapping patterns and improves the stability of hierarchical modeling.

\subsubsection{Magnitude Regularization}
Because the model learns residual affine parameters around the identity mapping, excessively large residuals may cause unstable enhancement and severe color shifts. We therefore regularize the overall magnitude of the bilateral grids as
\begin{equation}
\mathcal{L}_{mag}=\frac{1}{4}\sum_{k=0}^{3}\left\|\mathcal{G}^{(k)}\right\|_{1}.
\end{equation}
This term constrains the absolute magnitude of the residual affine coefficients, encouraging the network to learn moderate and stable corrections over the base RGB image rather than overly strong color perturbations.

\section{Experiments}

\subsection{Experimental Settings}

\subsubsection{Implementation Details}
Our method is implemented in PyTorch~\cite{NEURIPS2019_bdbca288}. During training, paired RAW-RGB samples are used for supervision. During inference, only RAW inputs are fed into the network. When the RAW image is stored as a Bayer mosaic, it is first packed into a four-channel representation according to the default RGGB pattern~\cite{Chen_2018_CVPR}; if the RAW input is already stored in a four-channel format, it is fed into the network directly. All input data are normalized to the range of $[0,1]$ after loading. During training, random cropping and basic geometric augmentation are applied. Specifically, the RGB crop size is set to $256\times256$, corresponding to a $128\times128$ packed RAW patch, and horizontal flipping, vertical flipping, and transposition are randomly used to improve generalization. During testing, no random cropping is used and inference is performed on full-resolution images. The network is trained with the AdamW optimizer~\cite{loshchilov2018decoupled} for 50 epochs. The initial learning rate is set to $2\times10^{-4}$ and the weight decay is set to $1\times10^{-4}$. The training batch size and test batch size are set to 8 and 1, respectively, with 4 data loading workers. The random seed is fixed to 3407. For the bilateral representation in RPBA-Net, the bilateral grid depth is set to 8, and the spatial resolutions of the four grids are set to 16, 32, 64, and 128, respectively. The scaling factor of the residual affine coefficients is set to 0.15. Mixed-precision training~\cite{micikevicius2018mixed} is enabled to improve computational efficiency.

\subsubsection{Datasets}
Experiments are conducted on two public benchmark datasets, namely ZRR~\cite{9022218} and MAI~\cite{Ignatov_2021_CVPR}.

\begin{itemize}
\item ZRR dataset~\cite{9022218}: It contains RAW images captured by Huawei P20 and sRGB images captured by a Canon camera. We follow the same training/testing split protocol as LiteISPNet~\cite{Zhang_2021_ICCV}, using 46800 RAW-sRGB image pairs for training and 1200 image pairs for testing.
\item MAI dataset~\cite{Ignatov_2021_CVPR}: It focuses on mapping RAW images captured by the Sony IMX586 sensor to the sRGB distribution of a Fujifilm camera. Since the official test split does not provide ground-truth sRGB images, we split the training set into a 9:1 ratio, resulting in 21700 pairs for training and 2400 pairs for evaluation.
\end{itemize}

It is worth noting that the RAW images in the ZRR dataset are 10-bit, whereas the RAW images in the MAI dataset are 12-bit.

\subsubsection{Evaluation Metrics}
To comprehensively evaluate reconstruction accuracy, perceptual quality, and deployment efficiency, we adopt full-reference metrics including PSNR, SSIM~\cite{1284395}, and LPIPS~\cite{Zhang_2018_CVPR}, as well as no-reference perceptual quality metrics including MUSIQ~\cite{Ke_2021_ICCV} and TOPIQ~\cite{10478301}. In addition, we report the number of model parameters (\#Params), floating-point operations (FLOPs), and inference time (Time). Specifically, SSIM~\cite{1284395} measures structural fidelity between the reconstructed image and the ground truth, while LPIPS~\cite{Zhang_2018_CVPR} reflects perceptual similarity. MUSIQ~\cite{Ke_2021_ICCV} and TOPIQ~\cite{10478301} are used to assess visual quality from a no-reference perspective. The \#Params, FLOPs, and Time metrics are further used to compare model size, computational complexity, and practical inference efficiency. In general, higher PSNR, SSIM, MUSIQ, and TOPIQ values and lower LPIPS values indicate better image quality, whereas smaller \#Params, FLOPs, and Time imply better deployment efficiency.

\subsection{Experimental Results}

\subsubsection{Compared Methods}
To comprehensively evaluate the effectiveness of RPBA-Net, we compare it with seven representative RAW-to-sRGB methods, including PyNet~\cite{Ignatov_2020_CVPR_Workshops}, AWNet-R~\cite{10.1007/978-3-030-67070-2_11}, AWNet-D~\cite{10.1007/978-3-030-67070-2_11}, MW-ISPNet~\cite{10.1007/978-3-030-67070-2_9}, LiteISPNet~\cite{Zhang_2021_ICCV}, FourierISP~\cite{He_2024_FourierISP}, and ISPDiffuser~\cite{DBLP:conf/aaai/RenJY0L25}. These baselines cover classical multi-scale ISP networks, adaptive weighting methods, lightweight architectures, frequency-domain decoupling methods, and diffusion-based methods, providing a broad comparison in terms of reconstruction quality, perceptual fidelity, and deployment efficiency.

\begin{table*}[!t]
\centering
\caption{Quantitative comparison results on the ZRR and MAI datasets.}
\label{tab:quantitative_comparison}
\footnotesize
\setlength{\tabcolsep}{4pt}
\renewcommand{\arraystretch}{1.15}
\resizebox{\textwidth}{!}{%
\begin{tabular}{l c ccccc ccccc ccccc}
\toprule
\multirow{2}{*}{Method}
& \multirow{2}{*}{\#Params$\downarrow$}
& \multicolumn{5}{c}{ZRR (Original GT)}
& \multicolumn{5}{c}{ZRR (Align GT with RAW)}
& \multicolumn{5}{c}{MAI} \\
\cmidrule(lr){3-7} \cmidrule(lr){8-12} \cmidrule(lr){13-17}
& & PSNR$\uparrow$ & SSIM$\uparrow$ & LPIPS$\downarrow$ & FLOPs$\downarrow$ & Time$\downarrow$
& PSNR$\uparrow$ & SSIM$\uparrow$ & LPIPS$\downarrow$ & FLOPs$\downarrow$ & Time$\downarrow$
& PSNR$\uparrow$ & SSIM$\uparrow$ & LPIPS$\downarrow$ & FLOPs$\downarrow$ & Time$\downarrow$ \\
\midrule
PyNet~\cite{Ignatov_2020_CVPR_Workshops}            & 47.549M & 21.19 & 0.747 & 0.193 & 672.337G & 24.00 ms  & 22.73 & 0.845 & 0.152 & 672.337G & 24.00 ms  & 23.81 & 0.848 & 0.139 & 219.539G & 17.32 ms \\
AWNet-R~\cite{10.1007/978-3-030-67070-2_11}          & 50.619M & 21.42 & 0.748 & 0.198 & \second{132.726G} & 26.82 ms  & 23.27 & 0.854 & 0.151 & \second{132.726G} & 26.82 ms  & 24.53 & 0.872 & 0.136 & \second{43.346G} & 19.90 ms \\
AWNet-D~\cite{10.1007/978-3-030-67070-2_11}          & 48.543M & 21.53 & 0.749 & 0.212 & 515.141G & 53.54 ms  & 23.38 & 0.850 & 0.164 & 515.141G & 53.54 ms  & 24.64 & 0.866 & 0.147 & 168.217G & 25.79 ms \\
MW-ISPNet~\cite{10.1007/978-3-030-67070-2_9}        & 29.219M & 21.42 & 0.754 & 0.213 & 177.337G & 23.98 ms  & 23.07 & 0.848 & 0.165 & 177.337G & 23.98 ms  & 25.02 & 0.885 & 0.133 & 57.906G  & 23.74 ms \\
LiteISPNet~\cite{Zhang_2021_ICCV}       & 9.048M  & 21.55 & 0.749 & 0.187 & 146.798G & \second{9.75 ms} & 23.76 & 0.873 & 0.133 & 146.798G & \second{9.75 ms} & 24.90 & 0.877 & 0.123 & 47.934G  & \second{7.63 ms} \\
FourierISP~\cite{He_2024_FourierISP}       & \second{7.590M} & 21.65 & \second{0.755} & 0.182 & 204.672G & 22.55 ms  & 23.93 & 0.874 & 0.124 & 204.672G & 22.55 ms  & 25.37 & 0.891 & 0.072 & 66.857G  & 13.80 ms \\
ISPDiffuser~\cite{DBLP:conf/aaai/RenJY0L25}     & 20.939M & \second{21.77} & 0.754 & \second{0.157} & 587.355G & 244.54 ms & \second{24.09} & \second{0.881} & \second{0.111} & 587.355G & 244.54 ms & \second{25.64} & \second{0.894} & \second{0.071} & 515.141G & 53.54 ms \\
RPBA-Net (Ours)  & \best{1.324M} & \best{22.01} & \best{0.783} & \best{0.151} & \best{11.395G} & \best{4.05 ms} & \best{24.32} & \best{0.891} & \best{0.107} & \best{11.395G} & \best{4.05 ms} & \best{26.78} & \best{0.899} & \best{0.068} & \best{5.238G} & \best{5.57 ms} \\
\bottomrule
\end{tabular}%
}
\end{table*}

\begin{table}[!t]
\centering
\caption{Comparison of no-reference perceptual quality metrics on the ZRR and MAI datasets.}
\label{tab:no_reference_comparison}
\footnotesize
\setlength{\tabcolsep}{6pt}
\renewcommand{\arraystretch}{1.15}
\resizebox{\columnwidth}{!}{%
\begin{tabular}{l cccc}
\toprule
\multirow{2}{*}{Method} & \multicolumn{2}{c}{ZRR} & \multicolumn{2}{c}{MAI} \\
\cmidrule(lr){2-3} \cmidrule(lr){4-5}
& MUSIQ$\uparrow$ & TOPIQ$\uparrow$ & MUSIQ$\uparrow$ & TOPIQ$\uparrow$ \\
\midrule
PyNet~\cite{Ignatov_2020_CVPR_Workshops}           & 43.796 & 0.362 & 39.823 & 0.445 \\
AWNet-R~\cite{10.1007/978-3-030-67070-2_11}         & 43.441 & 0.355 & 40.211 & 0.441 \\
AWNet-D~\cite{10.1007/978-3-030-67070-2_11}         & 45.100 & 0.362 & 39.839 & 0.432 \\
MW-ISPNet~\cite{10.1007/978-3-030-67070-2_9}       & 42.448 & 0.340 & 40.652 & 0.449 \\
LiteISPNet~\cite{Zhang_2021_ICCV}      & 47.310 & 0.370 & 40.365 & 0.445 \\
FourierISP~\cite{He_2024_FourierISP}      & 44.534 & 0.369 & 47.614 & \second{0.535} \\
ISPDiffuser~\cite{DBLP:conf/aaai/RenJY0L25}    & \second{50.117} & \second{0.392} & \second{48.032} & \second{0.535} \\
RPBA-Net (Ours) & \best{52.536} & \best{0.413} & \best{49.152} & \best{0.546} \\
\bottomrule
\end{tabular}%
}
\end{table}

\subsubsection{Quantitative Comparison}
We conduct a systematic quantitative comparison between the proposed method and the above competing methods on the ZRR~\cite{9022218} and MAI~\cite{Ignatov_2021_CVPR} datasets, and the results are reported in Tables~\ref{tab:quantitative_comparison} and~\ref{tab:no_reference_comparison}, respectively. Table~\ref{tab:quantitative_comparison} presents the full-reference metrics together with the model complexity and runtime results. It can be observed that RPBA-Net achieves the best PSNR, SSIM, and LPIPS results under all three evaluation settings, namely ZRR (Original GT), ZRR (Align GT with RAW)~\cite{Zhang_2021_ICCV}, and MAI. In particular, the PSNR values reach 22.01 dB, 24.32 dB, and 26.78 dB, respectively. Meanwhile, RPBA-Net also shows clear advantages in terms of parameter count, FLOPs, and inference time, with only 1.324M parameters, 11.395G FLOPs on ZRR, and 5.238G FLOPs on MAI. These results demonstrate that the proposed method not only improves RAW-to-RGB reconstruction quality, but also maintains high deployment efficiency with a lightweight architecture.

Furthermore, Table~\ref{tab:no_reference_comparison} reports the comparison results of two no-reference perceptual quality metrics, namely MUSIQ~\cite{Ke_2021_ICCV} and TOPIQ~\cite{10478301}. As can be seen, RPBA-Net achieves the best results for both metrics on both ZRR and MAI. Specifically, it reaches 52.536 and 0.413 on the ZRR dataset, and 49.152 and 0.546 on the MAI dataset, indicating that the proposed method can generate visually more pleasing results. Together with the full-reference results, these results show that RPBA-Net achieves a better overall balance among reconstruction accuracy, perceptual quality, and model efficiency.

\subsubsection{Qualitative Comparison}
\begin{figure*}[!t]
\centering
\includegraphics[width=\textwidth]{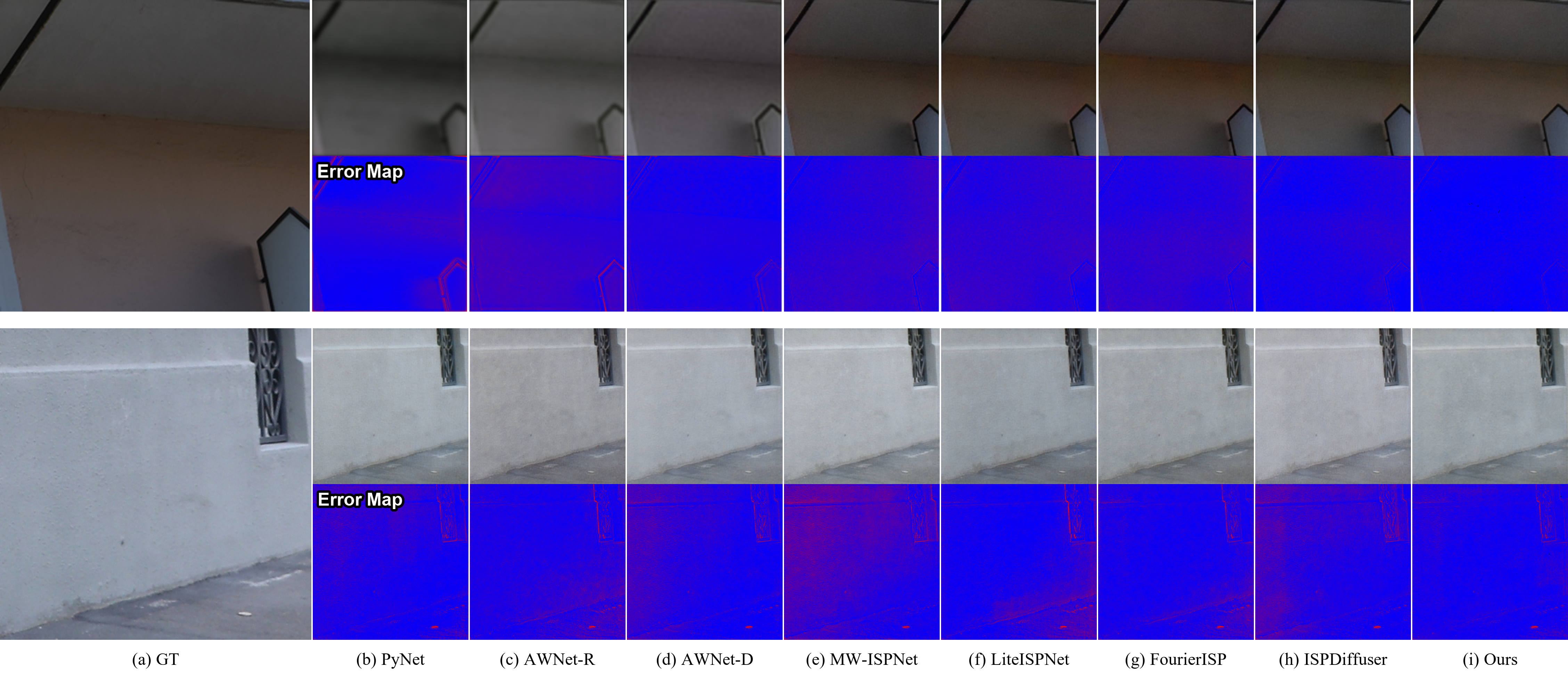}
\caption{Qualitative comparisons on MAI (top) and ZRR (bottom). RPBA-Net preserves clearer structures and yields lower error maps than competing methods.}
\label{fig:qualitative_comparison}
\end{figure*}

To further validate the visual superiority of the proposed method, we also conduct qualitative comparisons between RPBA-Net and the competing methods. As shown in Fig.~\ref{fig:qualitative_comparison}, the upper part presents the results on the MAI dataset and the lower part presents the results on the ZRR dataset. The competing methods suffer, to varying degrees, from insufficient texture recovery, blurred edges, local artifacts, and color shifts in complex scenes. In contrast, RPBA-Net delivers more stable texture detail recovery, edge structure preservation, and color reproduction, while producing error maps with lower residuals that are closer to the ground-truth references.

\subsection{Ablation Study}
To verify the effectiveness of each module in RPBA-Net, systematic ablation experiments are conducted on the MAI~\cite{Ignatov_2021_CVPR} dataset from four perspectives: residual affine base reconstruction, autoregressive adaptive bilateral slicing, pyramid bilateral affine grids with adaptive cross-layer fusion, and loss design.

\subsubsection{Ablation on Residual Affine Base Reconstruction}
\begin{figure}[!htbp]
\centering
\includegraphics[width=\columnwidth]{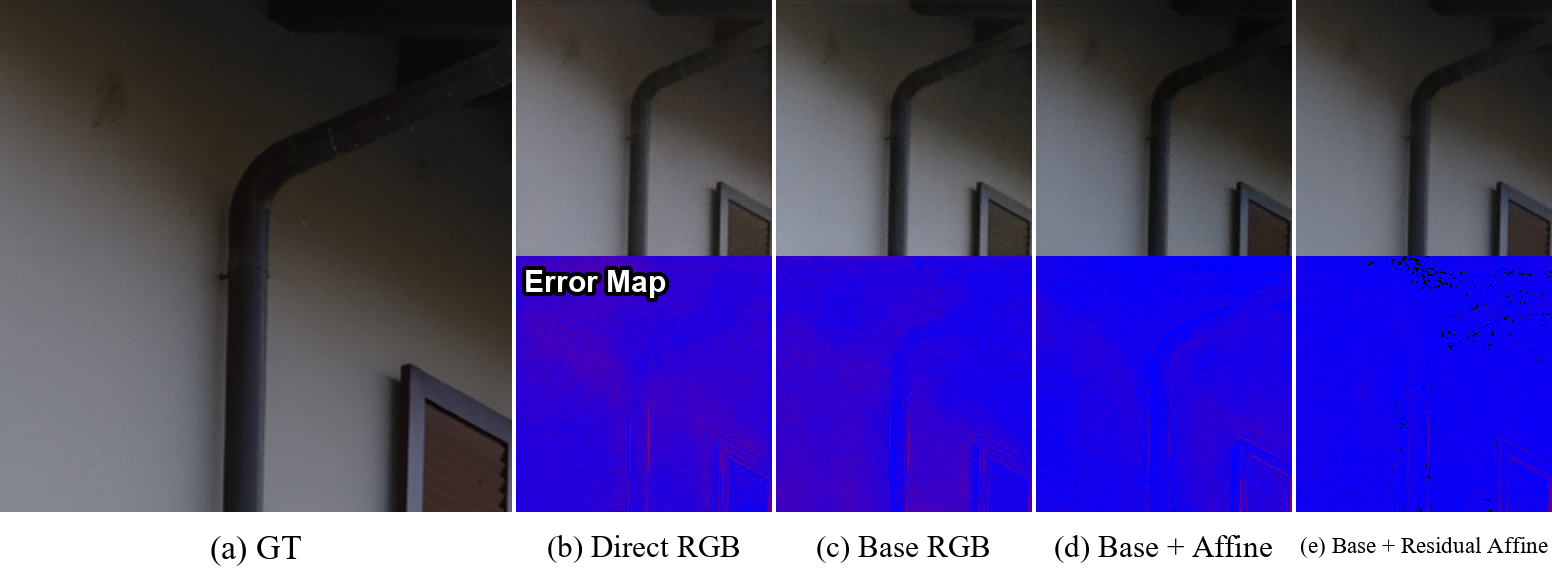}
\caption{Ablation on the residual affine design. Residual affine modeling improves color correction and reduces structural errors.}
\label{fig:ablation_base_affine_vis}
\end{figure}

Table~\ref{tab:ablation_base_affine} shows the effect of the base RGB estimation branch and residual affine design. Base RGB has limited restoration capacity. Although Direct RGB improves the reconstruction metrics, its reconstruction performance is still inferior to the affine-based designs. Base + Affine provides consistent gains, showing that affine correction effectively improves restoration quality. Base + Residual Affine achieves the best result, indicating that residual affine modeling under the identity-mapping prior yields better restoration performance while preserving mapping controllability. The corresponding visual comparison in Fig.~\ref{fig:ablation_base_affine_vis} further shows cleaner local structures and lower residual errors.

\begin{table}[!htbp]
\centering
\caption{Ablation results of residual affine base reconstruction on the MAI dataset.}
\label{tab:ablation_base_affine}
\footnotesize
\setlength{\tabcolsep}{6pt}
\renewcommand{\arraystretch}{1.12}
\resizebox{\columnwidth}{!}{%
\begin{tabular}{l cccccc}
\toprule
Variant & PSNR$\uparrow$ & SSIM$\uparrow$ & LPIPS$\downarrow$ & \#Params$\downarrow$ & FLOPs$\downarrow$ & Time$\downarrow$ \\
\midrule
Direct RGB & 24.52 & 0.877 & 0.101 & 1.276M & 4.880G & 5.21 ms \\
Base RGB & 24.03 & 0.866 & 0.116 & 1.102M & 4.619G & 5.04 ms \\
Base + Affine & 25.03 & 0.888 & 0.084 & 1.324M & 5.238G & 5.57 ms \\
Base + Residual Affine (Ours) & \best{26.78} & \best{0.899} & \best{0.068} & \best{1.324M} & \best{5.238G} & \best{5.57 ms} \\
\bottomrule
\end{tabular}%
}
\end{table}

\begin{figure}[!htbp]
\centering
\includegraphics[width=\columnwidth]{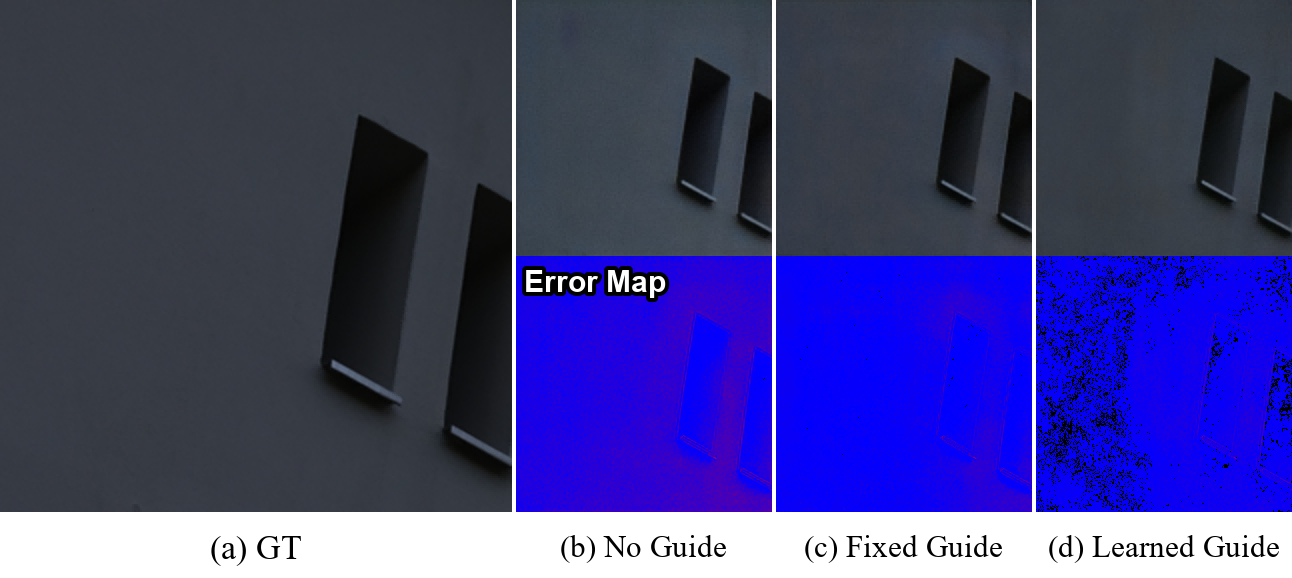}
\caption{Ablation on guide design. The learned guide enables more accurate content-aware coefficient selection.}
\label{fig:ablation_guide_design_vis}
\end{figure}

Furthermore, Table~\ref{tab:ablation_guide_design} reports the ablation results of guide design. No Guide achieves the worst reconstruction quality, indicating that the pixel-wise content awareness of the bilateral grids~\cite{10.1145/2980179.2982423,10.1145/3072959.3073592} is limited without guidance. Fixed Guide further improves the results, but its adaptivity remains limited. Learned Guide achieves the best performance, showing that a learned guide can more accurately assign local affine parameters to different pixels~\cite{10.1145/2980179.2982423,10.1145/3072959.3073592}. The qualitative results in Fig.~\ref{fig:ablation_guide_design_vis} also confirm that learned guidance improves local detail recovery and suppresses visible residual errors. Overall, these results demonstrate that base RGB estimation, residual affine correction, and learned guidance jointly support the effectiveness of residual affine base reconstruction.

\begin{table}[!htbp]
\centering
\caption{Ablation results of different guide designs on the MAI dataset.}
\label{tab:ablation_guide_design}
\footnotesize
\setlength{\tabcolsep}{6pt}
\renewcommand{\arraystretch}{1.12}
\resizebox{\columnwidth}{!}{%
\begin{tabular}{l cccccc}
\toprule
Variant & PSNR$\uparrow$ & SSIM$\uparrow$ & LPIPS$\downarrow$ & \#Params$\downarrow$ & FLOPs$\downarrow$ & Time$\downarrow$ \\
\midrule
No Guide & 24.61 & 0.879 & 0.098 & 1.301M & 5.095G & 5.36 ms \\
Fixed Guide & 25.09 & 0.889 & 0.083 & 1.309M & 5.163G & 5.43 ms \\
Learned Guide (Ours) & \best{26.78} & \best{0.899} & \best{0.068} & \best{1.324M} & \best{5.238G} & \best{5.57 ms} \\
\bottomrule
\end{tabular}%
}
\end{table}

\subsubsection{Ablation on Autoregressive Adaptive Bilateral Slicing}
\begin{figure}[!htbp]
\centering
\includegraphics[width=\columnwidth]{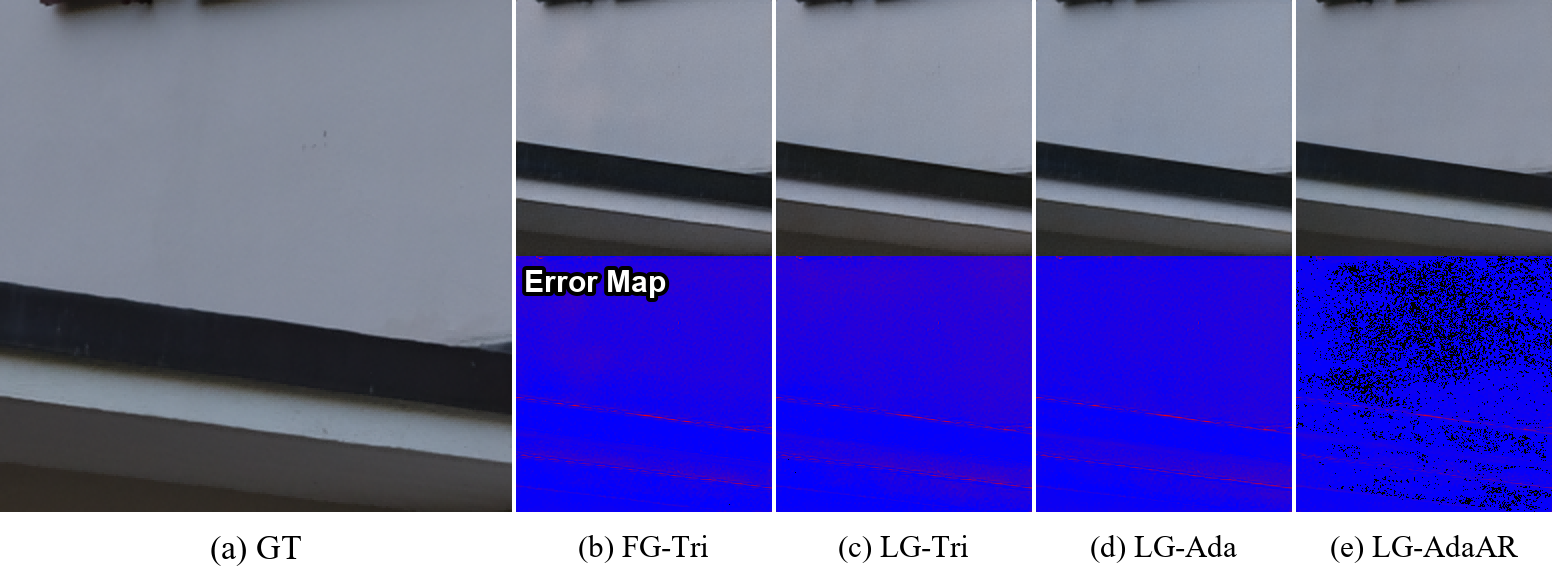}
\caption{Ablation on adaptive bilateral slicing. Autoregressive refinement improves detail recovery and suppresses artifacts.}
\label{fig:ablation_slice_vis}
\end{figure}

Table~\ref{tab:ablation_slice} reports the ablation results of autoregressive adaptive slicing variants. From FG-Tri to LG-Tri, PSNR improves by 0.57 dB and LPIPS decreases by 0.018, showing that learned guidance assigns bilateral coefficients more accurately for different pixels~\cite{10.1145/2980179.2982423,10.1145/3072959.3073592}. On this basis, LG-Ada further improves PSNR by 0.55 dB over LG-Tri and reduces LPIPS from 0.094 to 0.078, indicating that adaptive local interpolation models local content variation more effectively. With further coarse-to-fine autoregressive refinement, LG-AdaAR achieves the best reconstruction performance, with 26.78 dB PSNR, 0.899 SSIM, and 0.068 LPIPS. The visual comparison in Fig.~\ref{fig:ablation_slice_vis} further illustrates the progressive reduction of artifacts and error maps. Overall, from FG-Tri to LG-AdaAR, PSNR improves by 2.57 dB and LPIPS decreases by 0.044, with only a slight increase in FLOPs from 5.024G to 5.238G. These results show that learned guidance, adaptive local interpolation, and coarse-to-fine recursive refinement provide clear complementary benefits for bilateral slicing.

\begin{table}[!htbp]
\centering
\caption{Ablation results of autoregressive adaptive slicing variants on the MAI dataset.}
\label{tab:ablation_slice}
\footnotesize
\setlength{\tabcolsep}{6pt}
\renewcommand{\arraystretch}{1.12}
\resizebox{\columnwidth}{!}{%
\begin{tabular}{l cccccc}
\toprule
Variant & PSNR$\uparrow$ & SSIM$\uparrow$ & LPIPS$\downarrow$ & \#Params$\downarrow$ & FLOPs$\downarrow$ & Time$\downarrow$ \\
\midrule
FG-Tri & 24.21 & 0.870 & 0.112 & 1.286M & 5.024G & 5.22 ms \\
LG-Tri & 24.78 & 0.882 & 0.094 & 1.301M & 5.099G & 5.36 ms \\
LG-Ada & 25.33 & 0.892 & 0.078 & 1.316M & 5.183G & 5.48 ms \\
LG-AdaAR (Ours) & \best{26.78} & \best{0.899} & \best{0.068} & \best{1.324M} & \best{5.238G} & \best{5.57 ms} \\
\bottomrule
\end{tabular}%
}
\end{table}

\subsubsection{Ablation on Pyramid Bilateral Grids and Adaptive Fusion}
\begin{figure}[!htbp]
\centering
\includegraphics[width=\columnwidth]{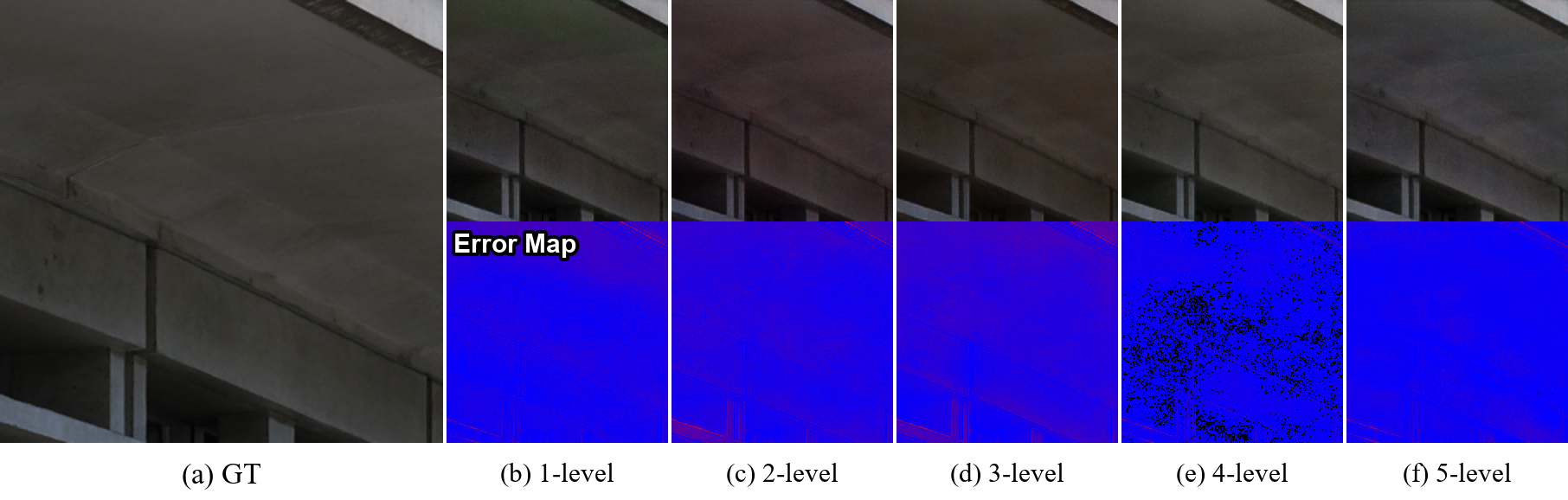}
\caption{Ablation on pyramid levels. The 4-level design better balances tone restoration and detail enhancement.}
\label{fig:ablation_levels_vis}
\end{figure}

This subsection analyzes the effectiveness of the structural design from three perspectives: the number of pyramid bilateral grid levels, residual affine scaling, and adaptive cross-layer fusion strategies. Table~\ref{tab:ablation_levels} studies the number of pyramid bilateral grid levels~\cite{10.1145/2980179.2982423,10.1145/3072959.3073592}. A single-level design is less effective at jointly handling global tone restoration and local texture enhancement. From 1 to 4 levels, the reconstruction performance improves consistently, showing that multi-level bilateral grids~\cite{10.1145/2980179.2982423,10.1145/3072959.3073592} enhance hierarchical coarse-to-fine modeling. The 4-level design achieves the best reconstruction result with 5.238G FLOPs, whereas the 5-level design raises FLOPs to 5.760G without providing further improvement. Fig.~\ref{fig:ablation_levels_vis} visually supports this observation, where the 4-level setting yields clearer structures and lower residual errors. This experiment indicates that the 4-level design achieves the best balance between reconstruction performance and computational complexity.

\begin{table}[!htbp]
\centering
\caption{Ablation results for different numbers of pyramid bilateral grid levels on the MAI dataset.}
\label{tab:ablation_levels}
\scriptsize
\setlength{\tabcolsep}{2pt}
\renewcommand{\arraystretch}{1.12}
\resizebox{\columnwidth}{!}{%
\begin{tabular}{l l cccccc}
\toprule
\# Levels & Spatial Resolutions & PSNR$\uparrow$ & SSIM$\uparrow$ & LPIPS$\downarrow$ & \#Params$\downarrow$ & FLOPs$\downarrow$ & Time$\downarrow$ \\
\midrule
1-level & 128 & 24.31 & 0.871 & 0.107 & 1.118M & 3.991G & 4.73 ms \\
2-level & 64, 128 & 24.82 & 0.882 & 0.092 & 1.186M & 4.434G & 5.04 ms \\
3-level & 32, 64, 128 & 25.29 & 0.891 & 0.079 & 1.255M & 4.803G & 5.31 ms \\
4-level (Ours) & 16, 32, 64, 128 & \best{26.78} & \best{0.899} & \best{0.068} & \best{1.324M} & \best{5.238G} & \best{5.57 ms} \\
5-level & 8, 16, 32, 64, 128 & 25.66 & 0.897 & 0.071 & 1.417M & 5.760G & 6.12 ms \\
\bottomrule
\end{tabular}%
}
\end{table}

\begin{figure}[!htbp]
\centering
\includegraphics[width=\columnwidth]{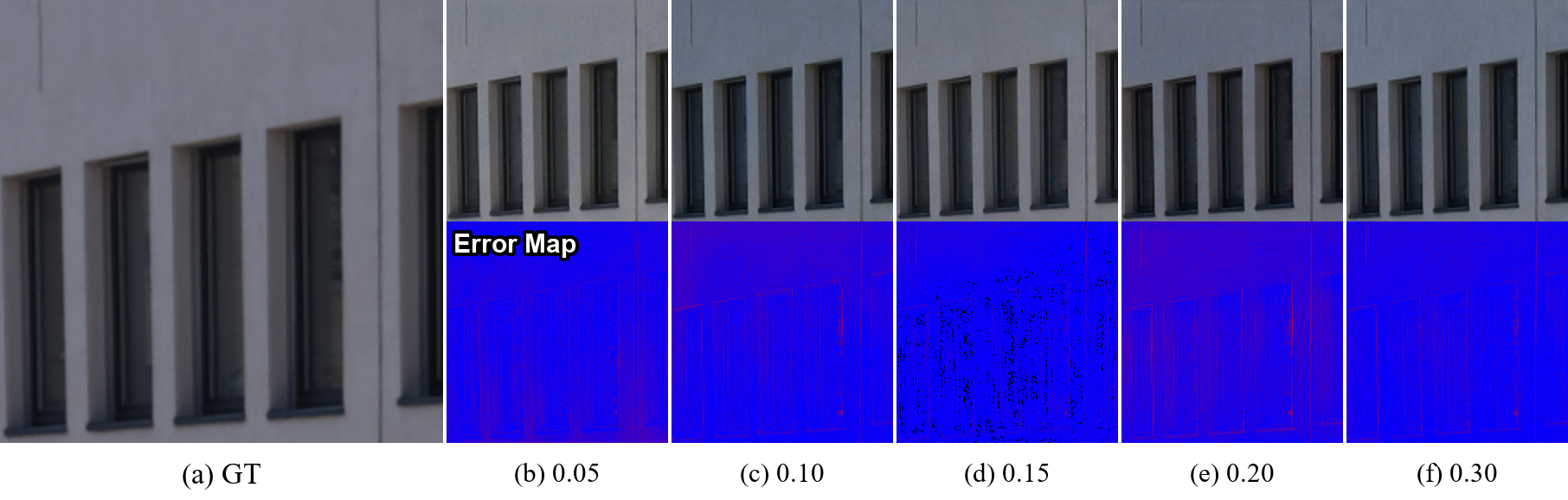}
\caption{Ablation on residual affine scaling. $\alpha=0.15$ provides stable correction while avoiding artifacts.}
\label{fig:ablation_alpha_vis}
\end{figure}

Table~\ref{tab:ablation_alpha} studies the residual scaling factor $\alpha$. A small scaling factor limits affine correction capability, whereas an excessively large scaling factor undermines overall stability. The best result is achieved at $\alpha = 0.15$. As shown in Fig.~\ref{fig:ablation_alpha_vis}, this setting also provides a visually better balance between detail correction and artifact suppression. These results indicate that a moderate residual scaling factor is crucial for balancing restoration capability and training stability.

\begin{table}[!htbp]
\centering
\caption{Ablation results for the scaling factor of the residual affine coefficients on the MAI dataset.}
\label{tab:ablation_alpha}
\footnotesize
\setlength{\tabcolsep}{8pt}
\renewcommand{\arraystretch}{1.12}
\begin{tabular}{c ccc}
\toprule
$\alpha$ & PSNR$\uparrow$ & SSIM$\uparrow$ & LPIPS$\downarrow$ \\
\midrule
0.05 & 24.77 & 0.881 & 0.094 \\
0.10 & 25.24 & 0.890 & 0.081 \\
0.15 (Ours) & \best{26.78} & \best{0.899} & \best{0.068} \\
0.20 & 25.41 & 0.893 & 0.078 \\
0.30 & 24.88 & 0.883 & 0.091 \\
\bottomrule
\end{tabular}%
\end{table}

\begin{figure}[!htbp]
\centering
\includegraphics[width=\columnwidth]{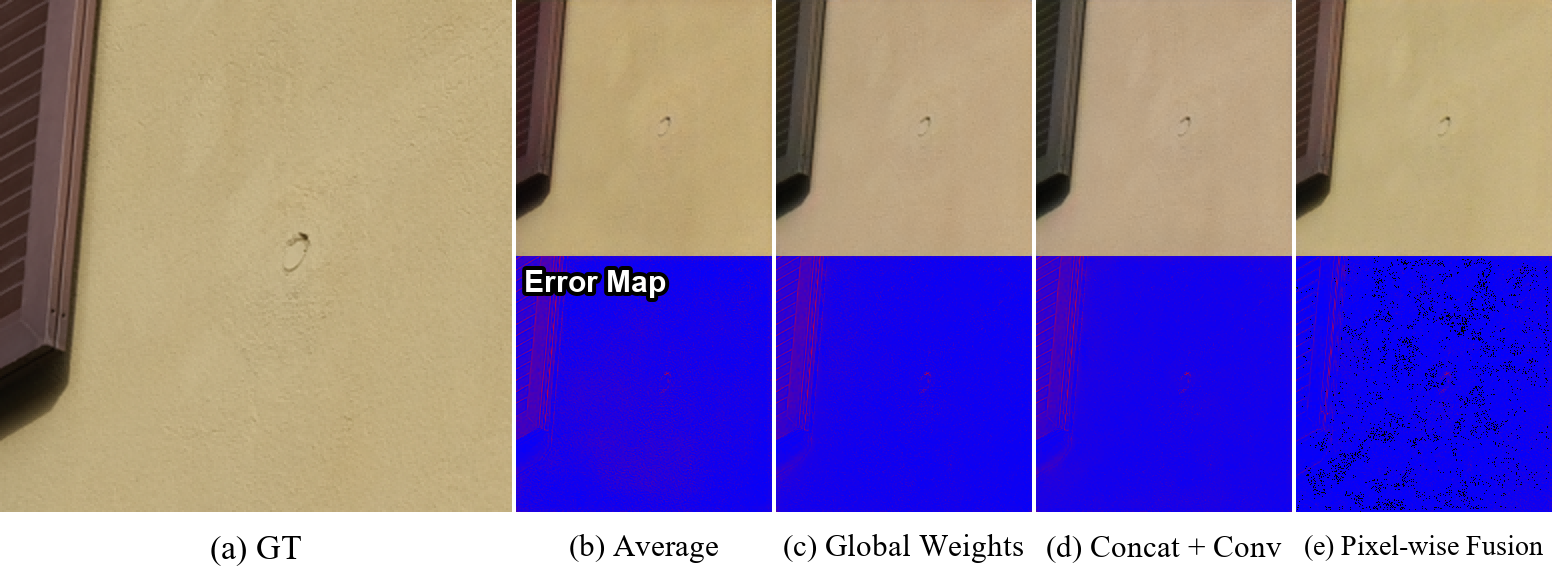}
\caption{Ablation on fusion strategy. Pixel-wise fusion better integrates multi-scale outputs and reduces residual errors.}
\label{fig:ablation_fusion_vis}
\end{figure}

Furthermore, Table~\ref{tab:ablation_fusion} compares different adaptive fusion strategies on the autoregressive adaptive slicing pipeline. Average fusion shows that simple averaging cannot model the varying contributions of different-scale branches across spatial regions. Global Weights improves the PSNR by 0.26 dB over Average, showing that learnable fusion is more effective than fixed averaging, although globally shared weights still limit content adaptivity. Concat + Conv further increases the PSNR to 25.38 dB, indicating the benefit of explicit cross-layer interaction. Pixel-wise Fusion achieves the best result, improving PSNR by 1.92 dB over Average and reducing LPIPS from 0.091 to 0.068 with only a marginal increase in FLOPs from 5.215G to 5.238G. The corresponding visualization in Fig.~\ref{fig:ablation_fusion_vis} shows that pixel-wise fusion better preserves local structures and suppresses residual artifacts. This result verifies the necessity of pixel-wise content-aware fusion after autoregressive adaptive slicing.

\begin{table}[!htbp]
\centering
\caption{Ablation results of different adaptive cross-layer fusion strategies on the MAI dataset.}
\label{tab:ablation_fusion}
\footnotesize
\setlength{\tabcolsep}{6pt}
\renewcommand{\arraystretch}{1.12}
\resizebox{\columnwidth}{!}{%
\begin{tabular}{l cccccc}
\toprule
Fusion Strategy & PSNR$\uparrow$ & SSIM$\uparrow$ & LPIPS$\downarrow$ & \#Params$\downarrow$ & FLOPs$\downarrow$ & Time$\downarrow$ \\
\midrule
Average & 24.86 & 0.884 & 0.091 & 1.312M & 5.215G & 5.49 ms \\
Global Weights & 25.12 & 0.889 & 0.083 & 1.313M & 5.216G & 5.50 ms \\
Concat + Conv & 25.38 & 0.894 & 0.076 & 1.336M & 5.264G & 5.63 ms \\
Pixel-wise Fusion (Ours) & \best{26.78} & \best{0.899} & \best{0.068} & \best{1.324M} & \best{5.238G} & \best{5.57 ms} \\
\bottomrule
\end{tabular}%
}
\end{table}

\subsubsection{Ablation on Loss Function Design}
\begin{figure}[!htbp]
\centering
\includegraphics[width=\columnwidth]{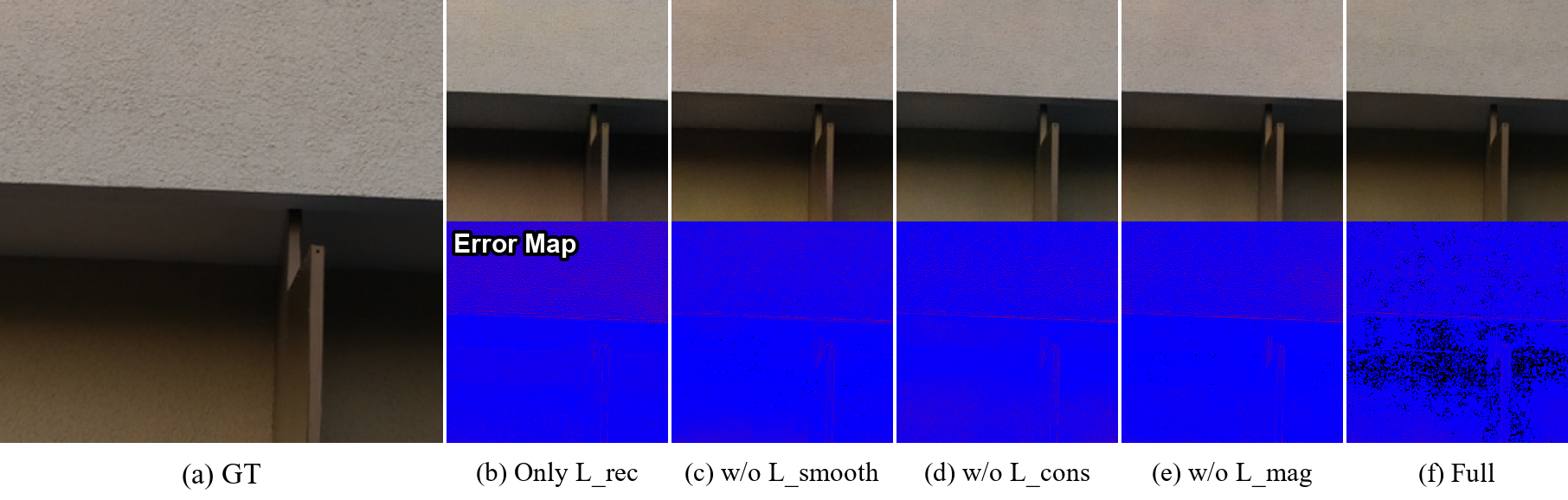}
\caption{Ablation on loss functions. The full objective improves visual fidelity and produces cleaner error maps.}
\label{fig:ablation_loss_vis}
\end{figure}

Table~\ref{tab:ablation_loss} reports the ablation results of the loss design. Using only $\mathcal{L}_{rec}$ clearly limits model performance, indicating that reconstruction supervision alone is insufficient to fully constrain bilateral-grid learning. Removing $\mathcal{L}_{smooth}$, $\mathcal{L}_{cons}$, or $\mathcal{L}_{mag}$ leads to varying degrees of performance degradation, showing that smoothness, cross-scale consistency, and magnitude constraints play important roles in local stability, scale coordination, and mapping controllability, respectively. Fig.~\ref{fig:ablation_loss_vis} further shows that the full objective produces clearer visual details and lower residual errors. The full objective achieves the best result, indicating that the three regularization terms work in a complementary manner.

\begin{table}[!htbp]
\centering
\caption{Ablation results of the loss design on the MAI dataset.}
\label{tab:ablation_loss}
\footnotesize
\setlength{\tabcolsep}{5pt}
\renewcommand{\arraystretch}{1.10}
\resizebox{\columnwidth}{!}{%
\begin{tabular}{l cccc ccc}
\toprule
\multirow{2}{*}{Variant} & \multicolumn{4}{c}{Loss Terms} & \multicolumn{3}{c}{Metrics} \\
\cmidrule(lr){2-5} \cmidrule(lr){6-8}
& $\mathcal{L}_{rec}$ & $\mathcal{L}_{smooth}$ & $\mathcal{L}_{cons}$ & $\mathcal{L}_{mag}$ & PSNR$\uparrow$ & SSIM$\uparrow$ & LPIPS$\downarrow$ \\
\midrule
Only $\mathcal{L}_{rec}$ & $\checkmark$ & $\times$ & $\times$ & $\times$ & 24.58 & 0.878 & 0.100 \\
w/o $\mathcal{L}_{smooth}$ & $\checkmark$ & $\times$ & $\checkmark$ & $\checkmark$ & 25.11 & 0.888 & 0.084 \\
w/o $\mathcal{L}_{cons}$ & $\checkmark$ & $\checkmark$ & $\times$ & $\checkmark$ & 25.22 & 0.890 & 0.081 \\
w/o $\mathcal{L}_{mag}$ & $\checkmark$ & $\checkmark$ & $\checkmark$ & $\times$ & 24.97 & 0.885 & 0.088 \\
Full (Ours) & $\checkmark$ & $\checkmark$ & $\checkmark$ & $\checkmark$ & \best{26.78} & \best{0.899} & \best{0.068} \\
\bottomrule
\end{tabular}%
}
\end{table}

\FloatBarrier

\section{Conclusion and Future Work}
This paper presents RPBA-Net, an interpretable residual pyramid bilateral affine network for RAW-domain ISP enhancement. Taking packed RAW as input, the proposed method achieves unified end-to-end modeling of demosaicing, color mapping, and detail enhancement through residual affine base reconstruction, pyramid bilateral affine grids with autoregressive adaptive slicing, and softmax-based adaptive cross-layer fusion. Meanwhile, grid smoothness, cross-scale consistency, and magnitude regularization are introduced to improve model stability, controllability, and structural interpretability. Experimental results show that RPBA-Net achieves better reconstruction performance than representative competing methods while maintaining low model complexity, and produces images with superior perceptual quality. Future work may further explore its extension to cross-device generalization, video RAW ISP, and efficient on-device deployment.

\bibliographystyle{IEEEbib}
\bibliography{bib/related_work}

\end{document}